\documentclass[sigconf]{acmart}

\usepackage{amsmath}
\usepackage{graphicx}
\usepackage{multirow}
\usepackage{multicol}
\usepackage{threeparttable}
\usepackage{enumitem}
\usepackage{subcaption}

\copyrightyear{2020}
\acmYear{2020}
\setcopyright{acmcopyright}\acmConference[CIKM '20]{Proceedings of the 29th ACM International Conference on Information and Knowledge Management}{October 19--23, 2020}{Virtual Event, Ireland}
\acmBooktitle{Proceedings of the 29th ACM International Conference on Information and Knowledge Management (CIKM '20), October 19--23, 2020, Virtual Event, Ireland}
\acmPrice{15.00}
\acmDOI{10.1145/3340531.3411949}
\acmISBN{978-1-4503-6859-9/20/10}

\begin{document}
\fancyhead{}
\title{Neural Logic Reasoning}
\author{Shaoyun Shi$^*$}
\thanks{* This work was conducted when Shaoyun Shi was visiting at Rutgers University. The first two authors contributed equally to the work.}
\affiliation{%
\institution{Tsinghua University}
\city{Beijing}
\country{China}
}
\email{shisy17@mails.tsinghua.edu.cn}

\author{Hanxiong Chen$^*$}
\affiliation{%
\institution{Rutgers University}
\city{New Brunswick}
\country{USA}
}
\email{hanxiong.chen@rutgers.edu}

\author{Weizhi Ma}
\affiliation{%
\institution{Tsinghua University}
\city{Beijing}
\country{China}
}
\email{mawz12@hotmail.com}

\author{Jiaxin Mao}
\affiliation{%
\institution{Tsinghua University}
\city{Beijing}
\country{China}
}
\email{maojiaxin@gmail.com}

\author{Min Zhang}
\affiliation{%
\institution{Tsinghua University}
\city{Beijing}
\country{China}
}
\email{z-m@tsinghua.edu.cn}

\author{Yongfeng Zhang}
\affiliation{%
\institution{Rutgers University}
\city{New Brunswick}
\country{USA}
}
\email{yongfeng.zhang@rutgers.edu}

\def\authors{Shaoyun Shi, Hanxiong Chen, Weizhi Ma, Jiaxin Mao, Min Zhang, Yongfeng Zhang}
\begin{abstract}
Recent years have witnessed the success of deep neural networks in many research areas. The fundamental idea behind the design of most neural networks is to learn similarity patterns from data for prediction and inference, which lacks the ability of cognitive reasoning. However, the concrete ability of reasoning is critical to many theoretical and practical problems. On the other hand, traditional symbolic reasoning methods do well in making logical inference, but they are mostly hard rule-based reasoning, which limits their generalization ability to different tasks since difference tasks may require different rules. Both reasoning and generalization ability are important for prediction tasks such as recommender systems, where reasoning provides strong connection between user history and target items for accurate prediction, and generalization helps the model to draw a robust user portrait over noisy inputs.

In this paper, we propose Logic-Integrated Neural Network (LINN) to integrate the power of deep learning and logic reasoning. LINN is a dynamic neural architecture that builds the computational graph according to input logical expressions. It learns basic logical operations such as AND, OR, NOT as neural modules, and conducts propositional logical reasoning through the network for inference. Experiments on theoretical task show that LINN achieves significant performance on solving logical equations and variables. Furthermore, we test our approach on the practical task of recommendation by formulating the task into a logical inference problem. Experiments show that LINN significantly outperforms state-of-the-art recommendation models in Top-K recommendation, which verifies the potential of LINN in practice.
\end{abstract}


\maketitle

\section{Introduction}

Deep neural networks have shown remarkable success in many fields, such as computer vision, natural language processing, information retrieval, and data mining. 
The design philosophy of most neural network architectures is learning statistical similarity patterns from large scale training data. For example, representation learning approaches learn vector representations from image or text for prediction, while metric learning approaches learn similarity functions for matching and inference. 

Though neural networks usually have good generalization ability on dataset with similar distribution, the design philosophy of these approaches makes it difficult for neural networks to conduct logical reasoning in many theoretical and practical problems. However, logical reasoning is an important ability for intelligence, and it is critical to many theoretical tasks such as solving logical equations, as well as practical tasks such as medical decision support systems, legal assistants, and personalized recommender systems. For example, in recommendation tasks, reasoning can help to model complex relationships between users and items (e.g., user likes item A but not item B $\rightarrow$ user likes item C), especially for those rare patterns, which is usually difficult for neural networks to capture.

One typical way to conduct reasoning is through logical inference. In fact, logical inference based on symbolic reasoning was the dominant approach to AI before the emerging of machine learning approaches, and it served as the underpinning of many expert systems in Good Old Fashioned AI (GOFAI). However, traditional symbolic reasoning methods for logical inference are mostly hard rule-based reasoning, which are not easily applicable to many real-world tasks due to the difficulty in defining the rules. For example, in personalized recommendation systems, if we consider each item as a symbol, we can hardly capture all of the relationships between the items based on manually designed rules. Besides, personalized user preferences bring various interacting sequences, which may result in conflicting reasoning rules, e.g., one user may like both item A and B, while another user who liked A may dislike B. Such noise in data makes it very challenging to design recommendation rules that properly generalize to new data. 


To unify the generalization ability of deep neural networks and logical reasoning, we propose Logic-Integrated Neural Network (LINN), a neural architecture to conduct logical inference based on neural networks. LINN adopts vectors to represent logic variables, and each basic logic operation (AND/OR/NOT) is learned as a neural module based on logic regularization. Since logic expressions that consist of the same set of variables may have completely different logical structures, capturing the structure information of logical expressions is critical to logical reasoning. To solve the problem, LINN dynamically constructs its neural architecture according to the input logical expression, which is different from many other neural networks with fixed computation graphs. By encoding logical structure information in neural architecture, LINN can flexibly process an exponential amount of logical expressions. Experiments on theoretical problems such as solving logical equations verified the superior logical inference ability of LINN compared to traditional neural networks. Extensive experiments on recommendation tasks show that by introducing logic constraints over the neural networks, LINN significantly outperforms state-of-the-art recommendation models.

The main contributions of this work are summarized as follows:
\begin{enumerate}
\item We propose a novel model named LINN, which has a modular design and empowers the neural networks with the ability of logical reasoning.
\item Experiments on theoretical task show that compared to traditional neural networks, LINN has significantly better ability on solving logical equations, which shows the great potential of LINN on theoretical tasks.
\item Experiments on practical task (recommendation) shows that LINN significantly outperforms state-of-the-art recommendation algorithms, which shows the great potential of LINN on practical tasks.
\end{enumerate}

The following part of the paper will include related work (Section \ref{sec:related}), proposed model (Section \ref{sec:nln}), theoretical experiments (Section \ref{sec:simulation}), practical experiments (Section \ref{sec:recommendation}), and conclusions (Section \ref{sec:conclusion}).


\section{Related Work}\label{sec:related}

\textbf{Logic and Symbolic AI.} The Good Old-Fashioned Artificial Intelligence (GOFAI)~\cite{haugeland1989artificial} was the dominant AI research paradigm from the mid-1950s to the late 1980s. It is based on the assumption that many aspects of intelligence can be achieved by the manipulation of symbols~\cite{newell1980physical}. Expert system is one of the most representative forms of logical/symbolic AI. Although their development is based on automatic rule mining, the generalization ability limits the application of these methods on many practical tasks~\cite{quinlan1991knowledge,galarraga2013amie}.

The problem is that not all of the practical problems (such as recommendation) are hard-rule logic reasoning problems.
For example, researchers have shown that logical equation solvers such as PicoSAT~\cite{biere2008picosat} performs well on clean logical equation problems, which include no logical contradictions in the data. However, it fails to give valid results on many practical tasks such as reasoning on visual images \cite{yi2018neural} and knowledge graphs \cite{hamilton2018embedding}. Similarly, personalized recommendation is a noisy reasoning task \cite{nguyen2007improving}, which required the ability of robust reasoning on data that includes contradictions, and it shows the importance of bridging the power of neural networks and logical reasoning for improved performance.

\noindent
\textbf{Deep Learning with Logic.} Deep learning has achieved great success in many areas. However, most of the existing methods are data-driven models that learn patterns from data without the ability of cognitive reasoning. Recently, several works used deep neural networks to solve logic problems. Hamilton et al.~\cite{hamilton2018embedding} embedded logical queries on knowledge graphs into vectors for knowledge reasoning. Johnson et al.~\cite{johnson2017inferring} and Yi et al.~\cite{yi2018neural} designed deep frameworks to generate programs and conduct visual reasoning. Yang et al.~\cite{yang2017differentiable} proposed a neural logic programming system to learn probabilistic first-order logical rules for knowledge base reasoning. Dong et al.~\cite{dong2019neural} proposed a neural logic machine architecture for relational reasoning and decision making. These works use pre-designed model structures to process different logical inputs, which is different from our LINN approach that constructs dynamic neural architectures. Although these works help in logical tasks, they are less flexible in terms of the model architecture, which makes them problem-specific and limits the application in a broader scope of theoretical and practical tasks.
Researchers are even trying to solve NP-complete Boolean satisfiability (SAT) problems with neural networks (NNs)~\cite{selsam2018learning}. It aims at deciding the satisfiability of a single logical expression using NNs, which shows that when properly designed, NNs have the ability to conduct symbolic reasoning on SAT problems. Our model also tackles NP-complete problems but goes even further to solve more challenging logical equation systems.

\noindent
\textbf{Neural Symbolic Learning.} Neural symbolic learning has a long history in the context of machine learning research. McCulloch and Pitts \cite{mcculloch1943logical} proposed one of the first neural systems for Boolean logic in \citeyear{mcculloch1943logical}. Neural symbolic learning gained further research in the 1990s and early 2000s. For example, researchers developed logical programming systems to make logical inference~\cite{garcez1999connectionist,holldobler1994towards}, and proposed neural frameworks for knowledge representation and reasoning~\cite{cloete2000knowledge,browne2001connectionist}. They adopt meticulously designed neural architectures to achieve the ability of logical inference. Garcez et al.~\cite{garcez2008neural} proposed a symbolic learning framework for non-classical logic, abductive reasoning, and normative multi-agent systems. However, it focuses more on hard logic reasoning, which is short of learning representations and generalization ability compared with deep neural networks, thus not suitable for reasoning over large-scale, heterogeneous, and noisy data. Neural symbolic learning received less attention during the 2010s when deep neural networks achieved great success on many tasks \cite{lecun2015deep}. However, neural symbolic learning has regained importance recently due to the difficulty for deep neural networks to conduct reasoning \cite{bengio2019from}. As a result, we explore LINN as an extension of neural networks for logic reasoning in this work, and demonstrate how the framework can be used for noisy reasoning tasks such as recommendation.




\section{Logic-Integrated Neural Networks}
\label{sec:nln}
In this section, we will introduce our Logic-Integrate Neural Network (LINN) architecture. In LINN, each logic variable in the logic expression is represented as a vector embedding, and each basic logic operation (i.e., AND/OR/NOT) is learned as a neural module. Most neural networks are developed based on fixed neural architectures that are independent from the input, either manually designed or learned through neural architecture search. Differently, the computational graph of our LINN architecture is input-dependent and built dynamically according to the input logic expression. We further leverage logic regularizers over the neural modules to guarantee that each module conducts the expected logical operation.

\vspace{-1ex}
\subsection{Logic Operations as Neural Modules}
An expression of propositional logic consists of logic constants (\texttt{True} or \texttt{False}, noted as \textit{T} or \textit{F}), logic variables ($v$), and basic logic operations (negation $\neg$, conjunction $\wedge$, and disjunction $\vee$). In LINN, the logic operations are learned as three neural modules. ~\citeauthor{leshno1993multilayer}~\cite{leshno1993multilayer} proved that multi-layer feed-forward networks with non-polynomial activation function can approximate any function. This provides us theoretical support for leveraging neural modules to learn the logic operations. Similar to most neural models in which input variables are learned as vector representations, in our framework, \textit{T}, \textit{F} and all logic variables are represented as vectors with the same dimension. Formally, suppose we have a set of logic expressions $E=\{e_i\}_{i=1}^m$ and their values $Y=\{y_i\}_{i=1}^m$ (either \textit{T} or \textit{F}), and they are constructed by a set of variables $V=\{v_i\}_{i=1}^n$, where $|V|=n$ is the number of variables. An example logic expression $e$ would look like $(v_i \wedge v_j)\vee \neg v_k=T$.

We use bold font to represent vectors, e.g. $\mathbf{v}_i$ is the vector representation of variable $v_i$, and $\mathbf{T}$ is the vector representation of logic constant \textit{T}, where the vector dimension is $d$. $\text{AND}(\cdot,\cdot)$, $\text{OR}(\cdot,\cdot)$, and $\text{NOT}(\cdot)$ are three neural modules. For example, $\text{AND}(\cdot,\cdot)$ takes two vectors $\mathbf{v}_i,\mathbf{v}_j$ as inputs, and the output $\mathbf{v}=\text{AND}(\mathbf{v}_i, \mathbf{v}_j)$ is the representation of $v_i\wedge v_j$, a vector of the same dimension $d$ as $\mathbf{v}_i$ and $\mathbf{v}_j$. The three modules can be implemented by various neural structures, as long as they are able to learn the logical operations.

\begin{figure}[t!]
\centering
\begin{subfigure}{0.23\textwidth}
\includegraphics[scale=0.45]{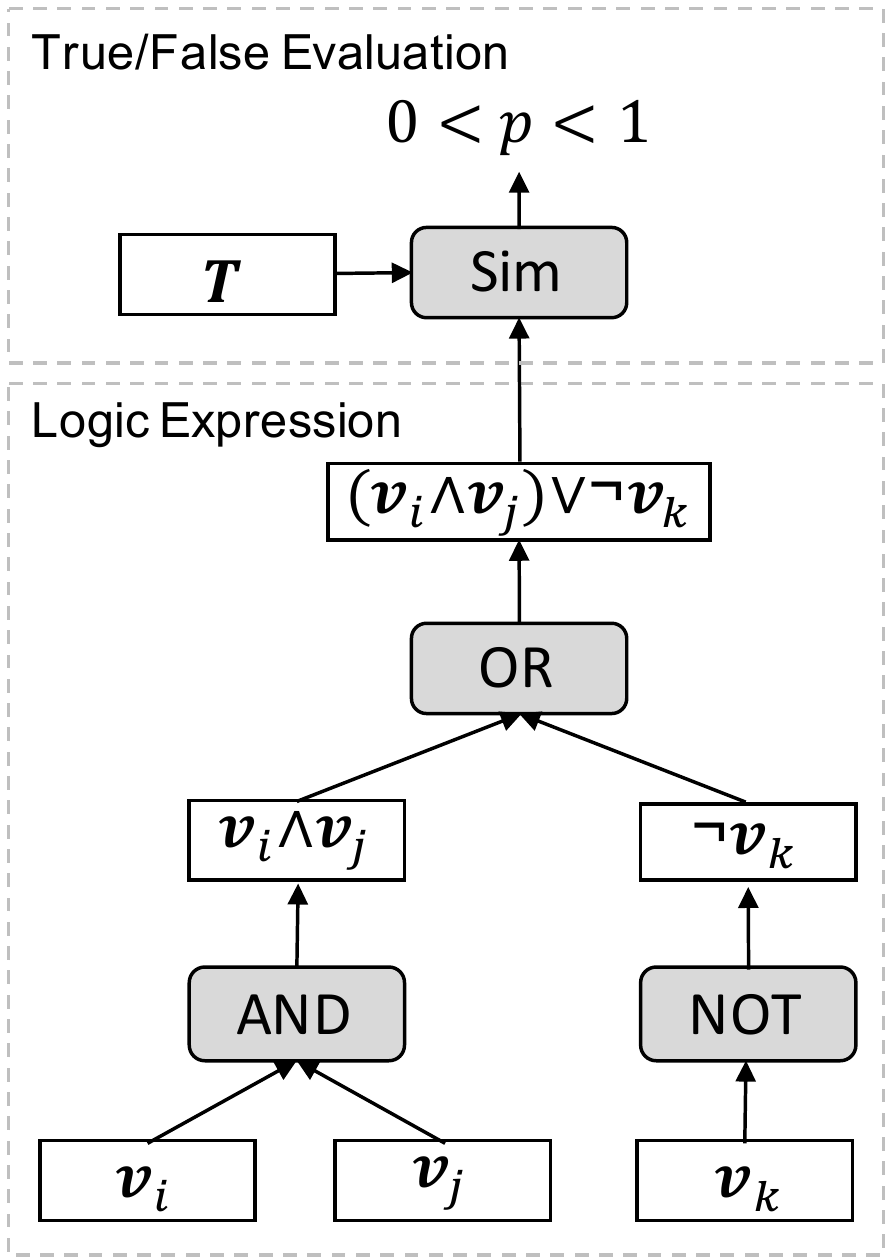}
\subcaption{
$(v_i\wedge v_j)\vee\neg v_k$}
\label{fig:example1}
\end{subfigure}
\begin{subfigure}{0.23\textwidth}
\includegraphics[scale=0.45]{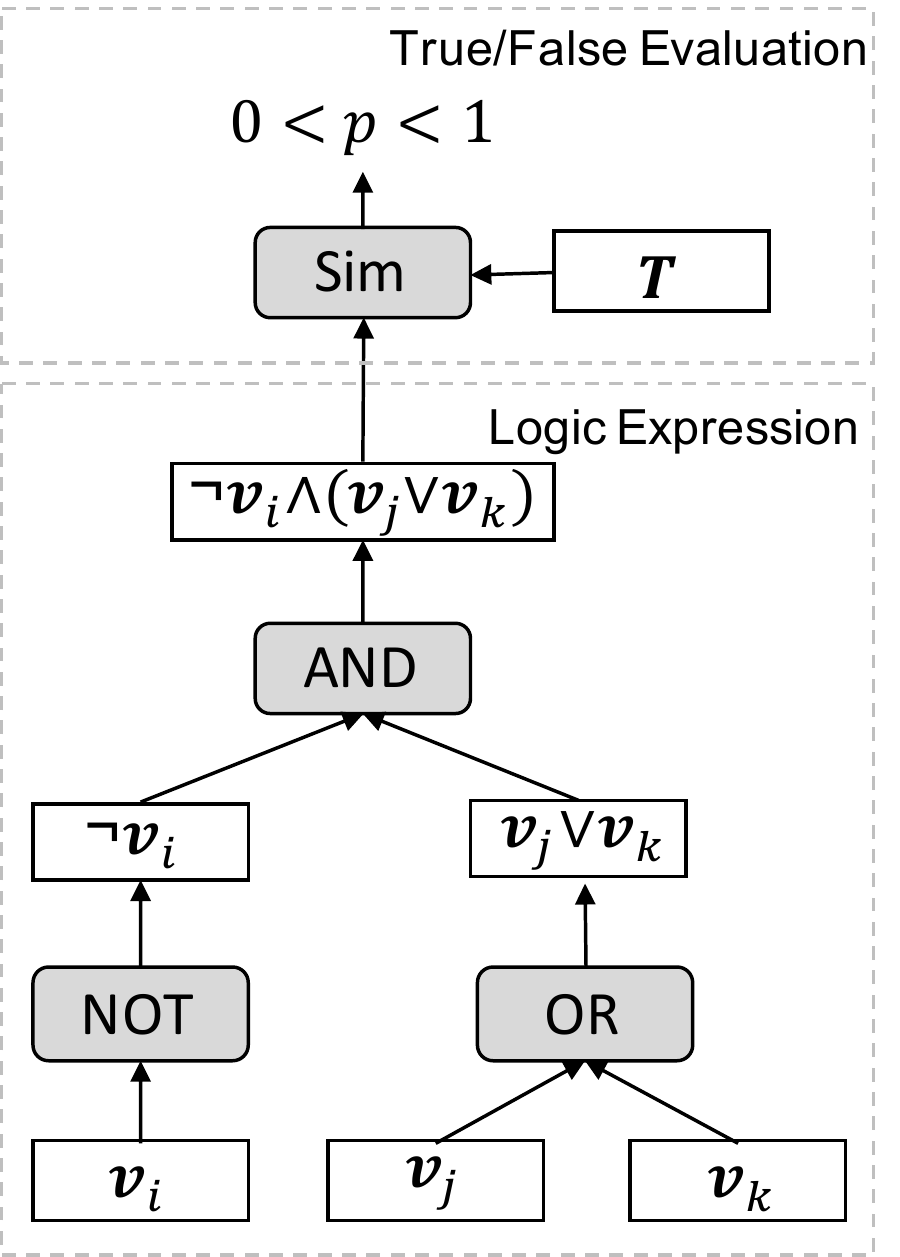}
\subcaption{
$\neg v_i\wedge (v_j\vee v_k)$
}
\label{fig:example2}
\end{subfigure}
\vspace{-0.1in}
\caption{Examples of the Logic-Integrated Neural Network (LINN) architecture. LINN learns each variable as a vector embedding, and learns each logic operation (i.e., $\wedge,\vee,\neg$) as a logic regularized neural module (i.e., AND, OR, NOT). For a given logic expression, e.g., $(v_i\wedge v_j)\vee\neg v_k$ in the left figure, LINN dynamically assembles the neural architecture according to the logic expression so as to learn the embedding of the whole expression. It then compares the expression with the constant True vector to evaluate the expression. For different input expressions, e.g., $\neg v_i\wedge (v_j\vee v_k)$ in the right figure, LINN reuses the variable embeddings and logic modules to assemble the architecture. In this way, LINN has the ability to model a compositional number of logic expressions.}
\vspace{-10pt}
\label{pic:framework}
\end{figure}

Figure~\ref{pic:framework}(a) is an example of the LINN architecture corresponding to the expression $(v_i \wedge v_j)\vee \neg v_k$. The lower box shows how the framework constructs a logic expression. Each intermediate vector represents part of the logic expression, and finally, we have the vector representation of the whole logic expression $\mathbf{e}=(\mathbf{v}_i \wedge \mathbf{v}_j)\vee \neg \mathbf{v}_k$. In this way, although the size of the variable embedding matrix grows linearly with the total number of logic variables, the total number of parameters in the operator networks keeps the same. Besides, since the expressions are organized as tree structures, we can calculate it recursively from leaves to the root. 

To evaluate the \textit{T}/\textit{F} value of the expression, we calculate the similarity between the expression vector and the $\mathbf{T}$ vector, as shown in the upper box, where \textit{T}, \textit{F} are short for logic constants \texttt{True} and \texttt{False}, respectively, and $\mathbf{T}$, $\mathbf{F}$ are their vector representations. Here $Sim(\cdot, \cdot)$ is also a neural module to calculate the similarity between two vectors and output a similarity value between 0 and 1. The output $p=Sim(\mathbf{e}, \mathbf{T})$ evaluates how likely LINN considers the expression to be true.

LINN can be trained with different task-specific loss functions for different tasks. For example,
training LINN on a set of expressions and predicting \textit{T}/\textit{F} values of other expressions can be considered as a classification problem, and we can adopt cross-entropy loss:
\begin{equation}
\label{eq:loss_ce}
L_{t} = L_{ce} = -\sum_{e_i \in E} y_i \log(p_i) + (1-y_i)\log(1-p_i)
\end{equation}
and for the top-n recommendation task, a pair-wise training loss based on Bayesian Personalized Ranking (BPR) \cite{rendle2009bpr} can be used:
\begin{equation}
\label{eq:loss_bpr}
L_{t} = L_{bpr} = -\sum_{e^+} \log\big(sigmoid(p(e^+) - p(e^-))\big)
\end{equation}
where $e_+$ are the expressions corresponding to the positive interactions in the dataset, and $e^-$ are the sampled negative expressions.

\begin{table*}[htbp]
    \vspace{-3mm}
    \caption{Logical regularizers and the corresponding logical rules}
    \vspace{-3mm}
    \label{tb:laws}
    \centering
    \begin{tabular}{p{1.5cm}p{3cm}p{2cm}p{6cm}}
      \toprule
      & Logical Rule     & Equation & Logic Regularizer $r_i$\\
      \midrule
      \multirow{2}{*}{NOT} & Negation & $\neg T = F$ & $r_1=\sum_{w\in W \cup \{T\}} Sim(\text{NOT}(\mathbf{w}),\mathbf{w})$    \\ 
      & Double Negation & $\neg (\neg w) = w$ & $r_2=\sum_{w\in W} 1-Sim(\text{NOT}(\text{NOT}(\mathbf{w})),\mathbf{w})$    \\ 
      \midrule
      \multirow{4}{*}{AND} & Identity & $w \wedge T = w$ & $r_3=\sum_{w\in W} 1-Sim(\text{AND}(\mathbf{w}, \mathbf{T}),\mathbf{w})$  \\
      & Annihilator & $w \wedge F = F$  & $r_4=\sum_{w\in W} 1-Sim(\text{AND}(\mathbf{w}, \mathbf{F}),\mathbf{F})$   \\
      & Idempotence & $w \wedge w = w$  &  $r_5=\sum_{w\in W} 1-Sim(\text{AND}(\mathbf{w}, \mathbf{w}),\mathbf{w})$ \\
      & Complementation & $w \wedge \neg w = F$  & $r_6=\sum_{w\in W} 1-Sim(\text{AND}(\mathbf{w}, \text{NOT}(\mathbf{w})),\mathbf{F})$   \\
      \midrule
      \multirow{4}{*}{OR} & Identity & $w \vee F = w$ & $r_7=\sum_{w\in W} 1-Sim(\text{OR}(\mathbf{w}, \mathbf{F}),\mathbf{w})$  \\
      & Annihilator & $w \vee T = T$  & $r_8=\sum_{w\in W} 1-Sim(\text{OR}(\mathbf{w}, \mathbf{T}),\mathbf{T})$  \\
      & Idempotence & $w \vee w = w$  & $r_9=\sum_{w\in W} 1-Sim(\text{OR}(\mathbf{w}, \mathbf{w}),\mathbf{w})$ \\
      & Complementation & $w \vee \neg w = T$ & $r_{10}=\sum_{w\in W} 1-Sim(\text{OR}(\mathbf{w}, \text{NOT}(\mathbf{w})),\mathbf{T})$  \\
      \bottomrule
    \end{tabular}
\vspace{-3mm}
\end{table*}

\subsection{\mbox{Logical Regularization over Neural Modules}}

So far, we only learned the logic operations $\text{AND}$, $\text{OR}$, $\text{NOT}$ as neural modules, but did not explicitly guarantee that these modules implement the expected logic operations. For example, any variable or expression $\mathbf{w}$ conjuncted with false should result in false, i.e. $\mathbf{w}\wedge\mathbf{F}=\mathbf{F}$, and a double negation should result in itself $\neg(\neg{\mathbf{w}})=\mathbf{w}$. Here we use $\mathbf{w}$ instead of $\mathbf{v}$ in the previous section, because $\mathbf{w}$ could either be a single variable (e.g., $\mathbf{v}_i$) or an expression in the middle of the calculation flow (e.g., $\mathbf{v}_i\wedge\mathbf{v}_j$).
A LINN that aims to implement logic operations should satisfy the basic logic rules. Although the neural architecture may implicitly learn the logic operations from data, it would be better if we apply constraints to guide the learning of logical operations. To achieve this goal, we define logic regularizers to regularize the behavior of the modules, so that they perform certain logical operations. A complete set of the logical regularizers are shown in Table~\ref{tb:laws}.

The regularizers are categorized by the three operations. 
To form the logical regularizers, these equations of laws are translated into the modules and variables in LINN, i.e. $r_i$'s in the table. It should be noted that these logical rules are not considered in the whole vector space $\mathbb{R}^d$, but in the vector space defined by LINN. Suppose the set of all input variables as well as intermediate states and final expressions observed during training process is represented as $W$, then only $\{\mathbf{w} | w\in W\}$ are constrained by the logical regularizers. Take Figure~\ref{pic:framework}(a) as an example, the corresponding $\mathbf{w}$ in Table~\ref{tb:laws} would include $\mathbf{v}_i$, $\mathbf{v}_j$, $\mathbf{v}_k$, $\mathbf{v}_i \wedge \mathbf{v}_j$, $\neg \mathbf{v}_k$ and $(\mathbf{v}_i \wedge \mathbf{v}_j) \vee \neg \mathbf{v}_k$. Logical regularizers encourage LINN to learn the neural module parameters to satisfy these laws over the variables/expressions involved in the model, which is much smaller than the whole vector space $\mathbb{R}^d$.

In LINN, the constant true vector $\mathbf{T}$ is randomly initialized and fixed during the training and testing process, which works as an anchor vector in the logic space that defines the true orientation. The false vector $\mathbf{F}$ is thus calculated with $\text{NOT}(\mathbf{T})$. 

Finally, logical regularizers ($R_{l}$) are added to the task-specific loss function $L_t$ with weight $\lambda_{l}$:
\begin{equation}\label{eq:logical_regularizer}
L_1 = L_t + \lambda_l R_{l} = L_t
 + \lambda_{l} \sum_i r_i
\end{equation}
where $r_i$ are the logic regularizers in Table~\ref{tb:laws}.


It should be noted that except for the logical regularizers listed above, a propositional logical system should also satisfy other logical rules such as the associativity, commutativity, and distributivity of AND/OR/NOT operations. For example, commutativity of the AND operation requires that $a\wedge b=b\wedge a$, while distributivity requires that $a\wedge (b\vee c)=(a\wedge b)\vee(a\wedge c)$. 

To consider the commutativity, the order of the variables joined by conjunctions or disjunctions is randomized when calculating the regularizers. For example, the network structure of $w \wedge T$ could be $\text{AND}(\mathbf{w}, \mathbf{T})$ or $\text{AND}(\mathbf{T}, \mathbf{w})$, and the network structure of $w \vee \neg w$ could be $\text{OR}(\mathbf{w}, \text{NOT}(\mathbf{w}))$ or $\text{OR}(\text{NOT}(\mathbf{w}), \mathbf{w})$. In this way, the model is encouraged to output the same vector representation when inputs are different forms of the same expression in terms of the commutativity. 

There is no explicit way to regularize the modules for other logical rules that correspond to more complex expression variants, such as distributivity and De Morgan laws\footnote{In propositional logic De Morgan laws state that $\neg(a\vee b)=\neg a \wedge \neg b$, while $\neg(a\wedge b)=\neg a \vee \neg b$}. To solve the problem, we make sure that the input expressions have the same normal form -- e.g., disjunctive normal form -- because any propositional logical expression can be transformed into a Disjunctive Normal Form (DNF) or Conjunctive Normal Form (CNF), and each expression corresponds to a computational graph. In this way, we can avoid the necessity to regularize the neural modules for distributivity and De Morgan laws.

\subsection{Length Regularization over Logic Variables}
We find that the vector length of logic variables, as well as intermediate or final logic expressions, may explode during the training process because simply increasing the vector length results in a trivial solution for optimizing Eq.\eqref{eq:logical_regularizer}. Constraining the vector length provides more stable performance, and thus a common $\ell_2$-length regularizer $R_{\ell}$ is added to the loss function with weight $\lambda_{\ell}$:
\begin{equation}
L_2 = L_t + \lambda_l R_{l} + \lambda_{\ell} R_{\ell} 
= L_t + \lambda_{l} \sum_i r_i  + \lambda_{\ell} \sum_{w\in W} \|\mathbf{w}\|_F^2
\end{equation}
Similar to the logical regularizers, $W$ here includes input variable vectors as well as all intermediate and final expression vectors.

Finally, we apply $\ell_2$-regularizer with weight $\lambda_{\Theta}$ to prevent the parameters from overfitting. Suppose $\Theta$ are all the model parameters, then the final loss function is:
\begin{equation}
\begin{aligned}
L &= L_t + \lambda_l R_{l} + \lambda_{\ell} R_{\ell} + \lambda_{\Theta}R_{\Theta}\\
  &= L_t + \lambda_{l} \sum_i r_i  + \lambda_{\ell} \sum_{w\in W} \|\mathbf{w}\|_F^2 + \lambda_{\Theta}\|\Theta\|_F^2
\end{aligned}
\end{equation}

\subsection{Implementation Details}

Our prototype task is defined in this way: given a number of training logical expressions and their \textit{T}/\textit{F} values, we train a LINN, and test if the model can solve the \textit{T}/\textit{F} value of the logic variables, and predict the value of new expressions constructed by the observed logic variables in training. In the following, we will first conduct experiments on a theoretical task to show that our LINN model has the ability to make propositional logical inference. Then, LINN is further applied to a practical recommendation problem to verify its performance in practical tasks.

We did not design fancy structures for different modules. Instead, some simple structures are effective enough to show the superiority of LINN. In our experiments, the $\text{AND}$ module is implemented by multi-layer perceptron (MLP) with one hidden layer:
\begin{equation}
    \text{AND}(\mathbf{w}_i, \mathbf{w}_j) = \mathbf{H}_{a2} f(\mathbf{H}_{a1} (\mathbf{w}_i\oplus\mathbf{w}_j) + \mathbf{b}_a)
\end{equation}
where $\mathbf{H}_{a1} \in \mathbb{R}^{d\times 2d}, \mathbf{H}_{a2} \in \mathbb{R}^{d\times d}, \mathbf{b}_a \in \mathbb{R}^d$ are the parameters of the $\text{AND}$ network. $\oplus$ means vector concatenation. $f(\cdot)$ is the activation function, and we use Rectified Linear Unit (ReLU) in our networks. The $\text{OR}$ module is built in the same way, and the $\text{NOT}$ module is similar but with only one vector as input:
\begin{equation}
    \text{NOT}(\mathbf{w}) = \mathbf{H}_{n2} f(\mathbf{H}_{n1} \mathbf{w} + \mathbf{b}_n)
\end{equation}
where $\mathbf{H}_{n1} \in \mathbb{R}^{d\times d}, \mathbf{H}_{n2} \in \mathbb{R}^{d\times d}, \mathbf{b}_n \in \mathbb{R}^d$ are the parameters of the $\text{NOT}$ network.

The similarity module is based on the cosine similarity of two vectors. To ensure that the output is formatted between 0 and 1, we scale the cosine similarity by multiplying a value $\alpha$, followed by a \textit{sigmoid} function:
\begin{equation}
    Sim(\mathbf{w}_i, \mathbf{w}_j) = sigmoid\left(\alpha \frac{\mathbf{w}_i \cdot \mathbf{w}_j}{\|\mathbf{w}_i\|\|\mathbf{w}_j\|}\right)
\end{equation}
The $\alpha$ is set to 10 in our experiments. We also tried other ways to calculate the similarity such as $sigmoid(\mathbf{w}_i \cdot \mathbf{w}_j)$ or MLP, and find that the above approach provides better performance.
\begin{table*}[htbp]
\vspace{-3mm}
      \caption{Performance on solving logical equations}
       \vspace{-3mm}
      \label{tab:performance_sim}
      \centering
      \begin{tabular}{p{2cm}|p{2.5cm}p{2.5cm}|p{2.5cm}p{2.5cm}}
        \toprule
        \multicolumn{1}{c}{}&
        \multicolumn{2}{c}{$n=1\times 10^3, m=3\times 10^3$}&
        \multicolumn{2}{c}{$n=1\times 10^4, m=3\times 10^4$}\\
        \cmidrule(lr){2-3} \cmidrule(lr){4-5}
        & Accuracy & RMSE & Accuracy & RMSE\\
        \midrule
        Bi-RNN~\cite{schuster1997BiRNN} &   0.6493 $\pm$ 0.0102   &   0.4736 $\pm$ 0.0032   
                                        &   0.6942 $\pm$ 0.0028   &   0.4492 $\pm$ 0.0009 \\
        Bi-LSTM~\cite{graves2005BiLSTM} &   0.5933 $\pm$ 0.0107   &   0.5181 $\pm$ 0.0162   
                                        &   0.6847 $\pm$ 0.0051   &   0.4494 $\pm$ 0.0020 \\
        CNN~\cite{kim2014CNN}           &   0.6380 $\pm$ 0.0043   &   0.5085 $\pm$ 0.0158   
                                        &   0.6787 $\pm$ 0.0025   &   0.4557 $\pm$ 0.0016 \\
        \midrule
        LINN-$R_l$                       &   \textit{0.8353 $\pm$ 0.0043}   &   \textit{0.3880 $\pm$ 0.0069} 
                                        &   \textit{0.9173 $\pm$ 0.0042}   &   \textit{0.2733 $\pm$ 0.0065}\\
        LINN                             &   \textbf{0.9440 $\pm$ 0.0064}*   &   \textbf{0.2318 $\pm$ 0.0124}*   
                                        &   \textbf{0.9559 $\pm$ 0.0006}*   &   \textbf{0.2081 $\pm$ 0.0018}*\\
      \bottomrule
    \end{tabular}
    \begin{minipage}{12cm}
      \begin{tablenotes}
      \item * Significantly better than the best of the other results (italic ones) with $p<0.05$
      \end{tablenotes}
    \end{minipage}
\vspace{-3mm}
\end{table*}

\section{Solving Logical Equations}
\label{sec:simulation}

Before applying LINN to practical tasks, in this section, we conduct experiments on a theoretical task based on simulated data to verify that our proposed LINN framework has the ability to conduct logic inference, which is difficult for traditional neural networks. Given a number of training logical expressions and their \textit{T}/\textit{F} values, we train a LINN, and test if the model can solve the \textit{T}/\textit{F} value of the logic variables, and predict the value of new expressions constructed by the observed logic variables.

We randomly generate $n$ variables $V=\{v_i\}$, and each variable is randomly assigned a value of either \textit{T} or \textit{F}. Then these variables are used to randomly create $m$ boolean expressions $E=\{e_i\}$ in disjunctive normal form (DNF) as the dataset. Each expression consists of 1 to 5 clauses separated by disjunction $\vee$, and each clause consists of 1 to 5 variables or the negation of variables connected by conjunction $\wedge$. We also conduct experiments on many other fixed or variational lengths of expressions, which have similar results. The \textit{T/F} values of the expressions $Y=\{y_i\}$ can be calculated according to the variables. But note that the \textit{T/F} values of the variables are invisible to the model. Here are some examples of the generated expressions when $n=100$:
\begin{equation*}
\begin{aligned}
    (\neg v_{80} \wedge v_{56} \wedge v_{71}) \vee (\neg v_{46} \wedge \neg v_{7} \wedge v_{51} \wedge \neg v_{47} \wedge v_{26})& \\
    \vee v_{45} \vee (v_{31} \wedge v_{15} \wedge v_{2} \wedge v_{46})& = T \\
    (\neg v_{19} \wedge \neg v_{65}) \vee (v_{65} \wedge \neg v_{24} \wedge v_{9} \wedge \neg v_{83})& \\
    \vee (\neg v_{48} \wedge \neg v_{9} \wedge \neg v_{51} \wedge v_{75})& = F \\
    \neg v_{98} \vee (\neg v_{76} \wedge v_{66} \wedge v_{13}) \vee (v_{97}  \wedge v_{89} \wedge v_{45} \wedge v_{83})& = T \\
    (v_{43} \wedge v_{21} \wedge \neg v_{53})& = F \\
\end{aligned}
\end{equation*}

We train the model based on a set of known DNFs and expect the model to predict the \textit{T/F} values for new DNFs, without the need to explicitly solve the \textit{T/F} value of each logical variable. It guarantees that there is an answer because the training equations are developed based on the same parameter set with known \textit{T/F} values. However, we do not use the variable \textit{T/F} information for model learning.

On simulated data, we use the Cross-Entropy loss $L_t$ (as in Equation~\ref{eq:loss_ce}). We randomly generate two datasets, of which one has $n=1\times 10^3$ variables and $m=3\times 10^3$ expressions, and the other is ten times larger. On the two datasets, weights of vector length regularizers $\lambda_{\ell}$ are set to 0.001, and weights of logic regularizers $\lambda_{l}$ are set to 0.01 and 0.1 respectively. Datasets are randomly split into the training (80\%), validation (10\%), and test (10\%) sets.

\subsection{Overall Performance}

The overall performances on test sets are shown on Table~\ref{tab:performance_sim}. \textbf{Bi-RNN} is bidirectional Vanilla RNN~\cite{schuster1997BiRNN} and \textbf{Bi-LSTM} is bidirectional LSTM~\cite{graves2005BiLSTM}. \textbf{CNN} is the Convolutional Neural Network~\cite{kim2014CNN}. They represent traditional neural networks, which regard logic expressions as sequences. \textbf{LINN-$R_l$} is the LINN model without logic regularizers (i.e., $\lambda_l=0$). We decide an equation as $T$ if the output similarity between the equation and $\textbf{T}$ vector is $\ge0.5$, and $F$ otherwise. Accuracy is the percentage of equations whose T/F value are correctly predicted. Root Mean Squre Error (RMSE) is calculated by considering the ground-truth value as 1 for $T$ and 0 for $F$.

The poor performance of Bi-RNN, Bi-LSTM, and CNN verifies that traditional neural networks, which ignore the logical structure of expressions, do not have the ability to conduct logical inference. Logical expressions are structural and have exponential combinations, which are difficult to learn by a fixed model architecture. We also see that Bi-RNN performs better than Bi-LSTM because the forget gate in LSTM may be harmful to model the variable sequence in expressions. LINN-$R_l$ provides a significant improvement over Bi-RNN, Bi-LSTM, and CNN because the structure information of the logical expressions is explicitly captured by the network structure. However, the behaviors of the logic modules in LINN-$R_l$ are freely trained without logical regularization. On this simulated data and many other problems requiring logical inference, logical rules are essential to model the internal relations. With the help of logic regularizers, the modules in LINN learn to perform expected logic operations, and finally, we can see that LINN achieves the best performance and significantly outperforms LINN-$R_l$.

\subsection{Weight of Logical Regularizers}

\begin{figure}[h]
  \vspace{-2mm}
  \centering
    \includegraphics[width=1.0\linewidth]{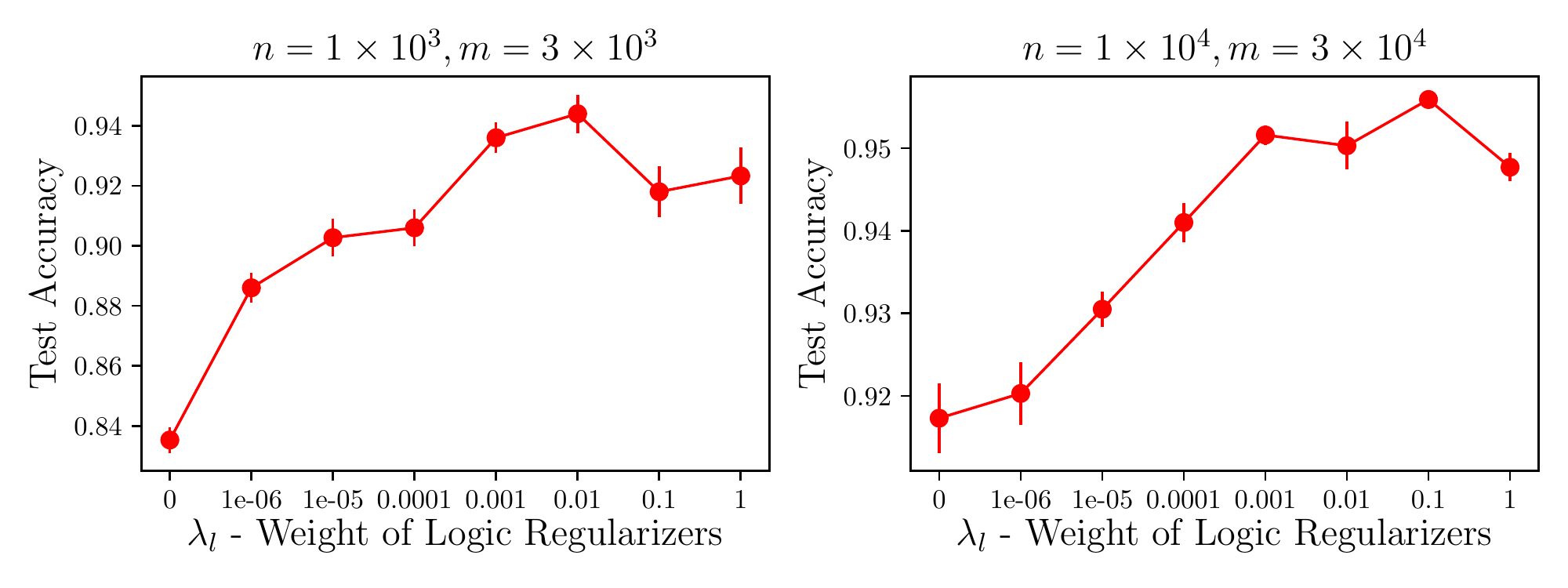}
    \vspace{-5mm}
    \caption{Performance under different choices of the logical regularization weight.}
    \label{fig:sim_logic_weights}
 \vspace{-2mm}
\end{figure}

To better understand the impact of logical regularizers, we test the model performance with different weights of logical regularizers. The results are shown in Figure~\ref{fig:sim_logic_weights}. When $\lambda_l=0$ (i.e., LINN-$R_l$), the performance is not so good, showing that structure information is not enough to inference the \textit{T/F} values of logic expressions. As $\lambda_l$ grows, the performance gets better, which shows that logical rules of the modules are essential for logical inference. However, a too-large $\lambda_l$ will result in a drop of performance, because the expressiveness power and generalization ability of the model may be significantly constrained by the logical regularizers, which force the model to performance hard-logic reasoning. In summary, proper weights of logic regularizers are helpful by unifying both the advantages of neural networks and logic reasoning.

\subsection{Solving Variables}

\begin{figure}[t!]
    \centering
    \includegraphics[width=1.0\linewidth,height=0.25\textheight]{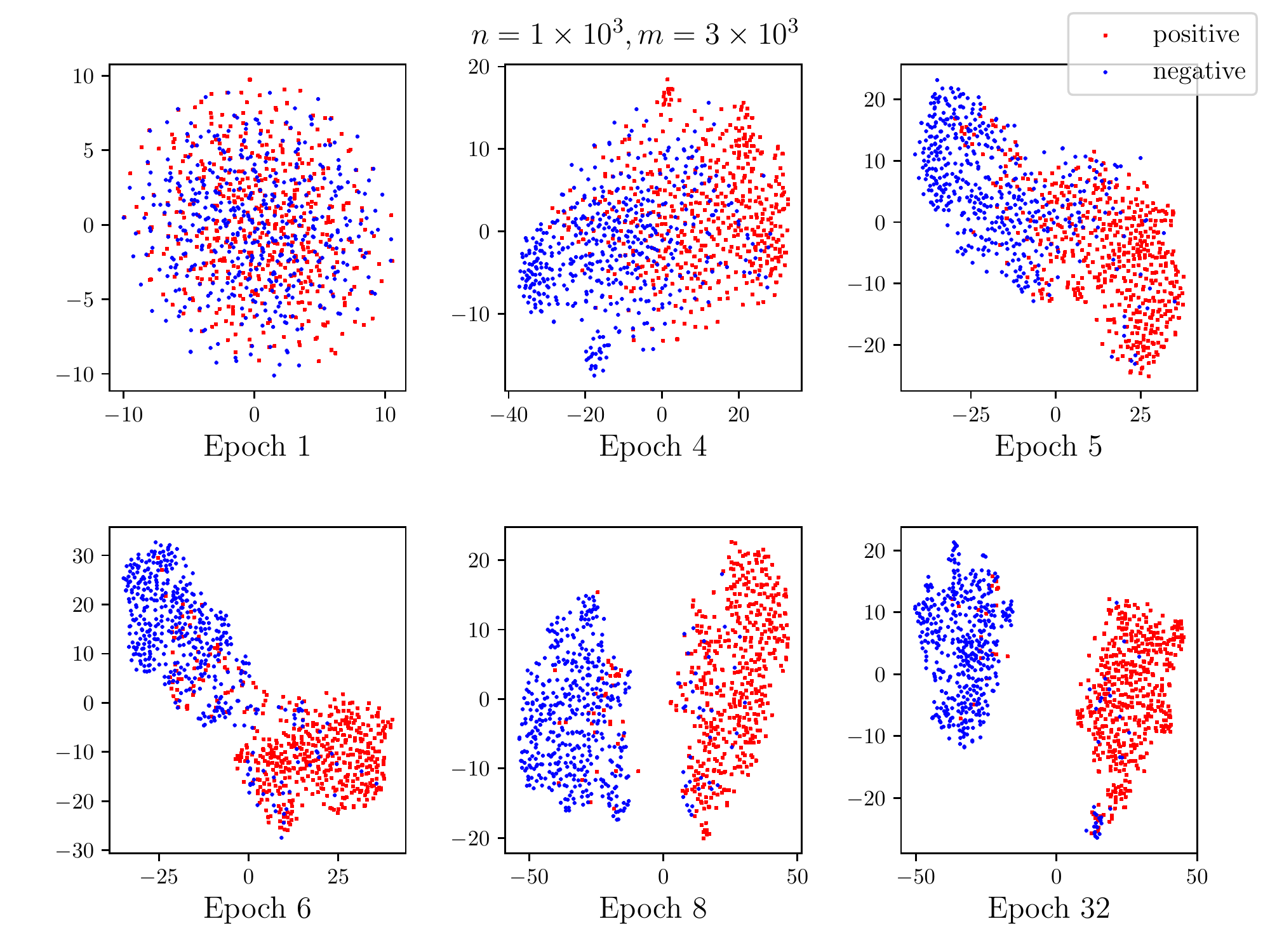}
    \vspace{-8mm}
    \caption{t-SNE visualization of how the variable embeddings change during model learning.}
    \label{fig:variable_tsne}
\vspace{-5mm}
\end{figure}

It would be interesting to see whether LINN can solve the \textit{T/F} value of the variables in logical equations. To do so, in different training epochs, we adopt t-SNE~\cite{maaten2008visualizing} to visualize the variable embeddings on a 2-D plot, shown in Figure~\ref{fig:variable_tsne}. We can see that at the beginning of the training process, the variable embeddings are randomly mixed together since they are randomly initialized. But they are gradually rearranged and finally clearly separated. We assign the variables in the two clusters as \textit{T} and \textit{F}, respectively. Compared with the ground-truth $T/F$ value of the variables, the accuracy is 95.9\%, which indicates a high accuracy of solving variables based on LINN. It means the model has the ability to inference the true/false values of the logic variables through a number of expressions, and thus it can further make reasoning on unseen logic expressions. Note that the model is not trained with the ground-truth \textit{T/F} values of the variables.

\section{Recommender Systems}
\label{sec:recommendation}

Experiments on simulated data verify that LINN is able to solve logic variables and make inferences on logical equations. It also reveals the prospect of LINN to apply on many other practical tasks as long as they can be formulated as logical expressions. Recommendation needs both the generalization ability of neural networks and reasoning ability of logic inference. As a result, we test the ability of LINN on making personalized recommendations.
To apply LINN into the recommendation task, we first convert it into a logically formulated problem. The key problem of recommendation is to understand user preferences according to historical interactions. Suppose there is a set of users $U=\{u_i\}$ and a set of items $V=\{v_j\}$, and the overall interaction matrix is $R=\{r_{i,j}\}_{|U|\times|V|}$. The interactions observed by the recommender system are the known values in matrix $R$. However, they are very sparse compared with the total size of the matrix $|U|\times|V|$. 
Recommendation models predict user's preference based on the user's history interactions (e.g., purchase history). Logic expressions can be easily used to model the item relationship in user history. For example, if a user bought an iPhone, he/she may need an iPhone case rather than an Android data line, i.e., $\text{\textit{iPhone}} \wedge \text{\textit{iPhone case}} = T$, while $\text{\textit{iPhone}} \wedge \text{\textit{Android data line}} = F$. Let $r_{i,j}=1/0$ if user $u_i$ likes/dislikes item $v_j$. Then if a user $u_i$ likes an item $v_{j_3}$ after a set of history interactions sorted by time $\{r_{i,j_1}=1, r_{i,j_2}=0\}$, the logical expression can be formulated as: 
\begin{equation}
\label{eq:recommend}
    (v_{j_1}\wedge v_{j_3}) \vee (\neg v_{j_2}\wedge v_{j_3}) \vee (v_{j_1} \wedge \neg v_{j_2}\wedge v_{j_3}) = T
\end{equation}
which means the user likes $v_{j_3}$ either because he/she likes  $v_{j_1}$, or because he/she dislikes $v_{j_2}$, or because he/she likes $v_{j_1}$ and dislikes $v_{j_2}$ at the same time. There are at least two advantages to take this format of expressions:
\begin{itemize}
    \item The model can filter out some unrelated interactions from user history. For example, if $v_{j_1}$, $v_{j_2}$, $v_{j_3}$ are iPhone, cat food and iPhone case, respectively, as long as $v_{j_1} \wedge v_{j_3}$ is close to the vector $\mathbf{T}$, it does not matter what the vector of $\neg v_{j_2} \wedge v_{j_3}$ look like. This characteristic encourages the model to find proper solutions to multiple expressions through collaborative filtering, and contribute to learning the internal relationship between items.
    \item After training the model, given the user history and a target item, the model not only predicts how much the user prefers the item, but also finds out which items are related by evaluating each conjunction term, i.e. $v_{j_1}\wedge v_{j_3}$, $\neg v_{j_2}\wedge v_{j_3}$ and $v_{j_1} \wedge \neg v_{j_2}\wedge v_{j_3}$ in Equation~\ref{eq:recommend}. If $v_{j_1} \wedge v_{j_3}$ is close to the vector $\mathbf{T}$, the recommender system can give a explanation that it recommends item $v_{j_3}$ to the user because he/she likes $v_{j_1}$. This makes a step to explainable recommendation \cite{zhang2020explainable,zhang2014explicit}, which is an important aspect in recommender system research.
\end{itemize}

In Equation~\ref{eq:recommend}, the first two conjunction terms are the first-order relationships between items, and the third one is a second-order relationship. If the user has more historical interactions, there can be higher-order and more relationships, but the number of disjunction terms in Equation \ref{eq:recommend} will explode exponentially. One solution is to sample terms randomly or using some designed sampling strategies. In this work, for model simplicity, we only consider the first-order relationships in experiments, which is good enough to surpass many state-of-the-art methods. Considering higher-order relationships can be the future improvements of the work. As far as we know, there are few works considering high-order recommendation rules or explanations for recommendation.

It should be noted that the above recommendation model is non-personalized, i.e., we do not learn user embeddings for prediction and recommendation, but instead only rely on item relationships for recommendation. However, as we will show in the following experiments, such a simple non-personalized model outperforms state-of-the-art personalized neural sequential recommendation models, which implies the great power and potential of reasoning in recommendation tasks. Extending the model to personalized versions will be considered in the future.

\subsection{Experimental Settings}

\subsubsection{Training and Evaluation}

The models are evaluated on the Top-K Recommendation task. We use the pair-wise learning strategy~\cite{rendle2009bpr} to train the model, which is a commonly used training strategy in many ranking tasks. In detail, we use the positive interactions in the datasets to train the baseline models and use the expressions corresponding to the positive interactions to train our LINN model. For each positive interaction $v^+$, we randomly sample an item the user dislikes or has never interacted with before as the negative sample $v^-$ in each epoch. Then the loss function of the base models is:
\begin{equation}
\label{eq:bpr_baseline}
   L = -\sum_{v^+} \log\big(sigmoid(p(v^+) - p(v^-))\big) + \lambda_\Theta \|\Theta\|_F^2
\end{equation}
where $p(v^+)$ and $p(v^-)$ are the predictions of $v^+$ and $v^-$, respectively, and $\lambda_\Theta \|\Theta\|_F^2$ is $\ell_2$-regularization. The loss function encourages the predictions of positive interactions to be higher than the negative samples. For our LINN, suppose the logic expression with $v^+$ as the target item is $e^+=(\cdot\wedge v^+)\vee\cdots\vee(\cdot\wedge v^+)$, then the negative expression is $e^-=(\cdot\wedge v^-)\vee\cdots\vee(\cdot\wedge v^-)$, which has the same history interactions as $e^+$ but replaces $v^+$ with $v^-$. Then the final loss function of our LINN model is:
\begin{equation}
\begin{aligned}
      L = &-\sum_{e^+} \log\big(sigmoid(p(e^+) - p(e^-))\big) \\
          &+ \lambda_{l} \sum_i r_i  + \lambda_{\ell} \sum_{w\in W} \|\mathbf{w}\|_F^2 + \lambda_{\Theta}\|\Theta\|_F^2
\end{aligned}
\end{equation}
where $p(e^+)$ and $p(e^-)$ are the predictions of $e^+$ and $e^-$, respectively, and other parts are the logic, vector length and $\ell_2$-regularizers as mentioned before. In Top-K evaluation, we sample 100 $v^-$ for each $v^+$ and evaluate the rank of $v^+$ in these 101 candidates, which is a common way in many previous works. Recommendation performance is evaluated on two metrics (mean of all evaluation samples):
\begin{itemize}
    \item \textbf{nDCG@K}: larger is better. It is one of the most popular ranking metrics in information retrieval, which is higher when the results are more similar to the ideal ranking. We consider the top-10 results, i.e., $K=10$.
    \item \textbf{Hit@K}: larger is better. It is the percentage of ranking lists that include at least one positive item in the top-K highest ranked items. We use $K=1$, which means that we evaluate whether the top-ranked item is positive.
\end{itemize}

\subsubsection{Datasets}
Experiments are conducted on two publicly available datasets:

$\bullet$ \textbf{ML-100k}~\cite{harper2016movielens}. It is maintained by Grouplens\footnote{\url{https://grouplens.org/datasets/movielens/100k/}}, which has been used by researchers for many years. It includes 100,000 ratings ranging from 1 to 5 from 943 users and 1,682 movies.

$\bullet$ \textbf{Amazon Electronics}~\cite{he2016ups}. Amazon Dataset\footnote{\url{http://jmcauley.ucsd.edu/data/amazon/index.html}} is a public e-commerce dataset. It contains reviews and ratings of items given by users on Amazon -- a popular e-commerce website. We use a subset in the area of Electronics, containing 1,689,188 ratings ranging from 1 to 5 from 192,403 users and 63,001 items, which is larger and much more sparse than the ML-100k dataset. 

\begin{table}[htbp]
 \vspace{-3mm}
  \caption{Statistics of the Datasets}
\vspace{-4mm}
  \label{tab:datasets}
  \begin{tabular}{ccccc}
    \toprule
    Dataset & \#User & \#Item & \#Positive & \#Negative \\
    \midrule
    ML-100k & 943 & 1,682 & 55,375 & 44,625 \\
    Electronics & 192,403 & 63,001 & 1,356,067 & 333,121 \\
  \bottomrule
\end{tabular}
\end{table}

The statics of two datasets are summarized in Table~\ref{tab:datasets}. The ratings are transformed into 0 and 1. Ratings equal to or higher than 4 ($r_{i,j}\geq 4$) are transformed to 1, which means positive attitudes (like). Other ratings ($r_{i,j}\le 3$) are converted to 0, which means negative attitudes (dislike). Then, each user's interactions are sorted by time, and each positive interaction of the user is translated to a logic expression in the way mentioned above (Equation \ref{eq:recommend}). We ensure that the expressions corresponding to a user's earliest 5 positive interactions are in the training set. For those users with no more than 5 positive interactions, all the expressions are in the training set. For the remaining data, the last two positive expressions of every user are distributed into the validation set and test set, respectively (test set is preferred if there remains only one expression for the user). All the other expressions are in the training set. This way of data partition and evaluation is usually called the Leave-One-Out setting in personalized recommendation research.

\begin{table*}[t!]
\vspace{-3mm}
  \caption{Performance on the recommendation task}
   \vspace{-3mm}
  \label{tab:performance_rec}
  \centering
  \begin{tabular}{p{2cm}|p{2.5cm}p{2.5cm}r|p{2.5cm}p{2.5cm}r}
    \toprule
    \multicolumn{1}{c}{}&
    \multicolumn{3}{c}{\textbf{ML-100k}}&
    \multicolumn{3}{c}{\textbf{Amazon Electronics}}\\
    \cmidrule(lr){2-4} \cmidrule(lr){5-7}
    & nDCG@10 & Hit@1 & time/epoch & nDCG@10 & Hit@1 & time/epoch\\
    \midrule
    BPRMF~\cite{rendle2009bpr}        &   0.3664 $\pm$ 0.0017   &   0.1537 $\pm$ 0.0036   &   4.9s
                                      &   0.3514 $\pm$ 0.0002   &   0.1951 $\pm$ 0.0004   &   112.1s\\
    SVD++~\cite{koren2008SVDPP}       &   0.3675 $\pm$ 0.0024   &   0.1556 $\pm$ 0.0044   &   30.4s
                                      &   0.3582 $\pm$ 0.0004   &   0.1930 $\pm$ 0.0006   &   469.3s\\
    STAMP~\cite{liu2018STAMP}         &   0.3943 $\pm$ 0.0016   &   0.1706 $\pm$ 0.0022   &  8.3s
                                      &   0.3954 $\pm$ 0.0003   &   0.2215 $\pm$ 0.0003   &  352.7s\\
    \midrule
    GRU4Rec~\cite{hidasi2015GRU4Rec}  &   0.3973 $\pm$ 0.0016   &   0.1745 $\pm$ 0.0038   &  7.1s
                                      &   0.4029 $\pm$ 0.0009   &   0.2262 $\pm$ 0.0009   &  225.0s\\
    NARM~\cite{li2017NARM}            &   \textit{0.4022 $\pm$ 0.0015}   &   0.1771 $\pm$ 0.0016  &  9.6s
                                      &   0.4051 $\pm$ 0.0006   &   0.2292 $\pm$ 0.0005  &  268.8s\\
    \midrule
    LINN-$R_l$                         &   0.4022 $\pm$ 0.0027   &   \textit{0.1783 $\pm$ 0.0043}   &   20.7s
                                      &   \textit{0.4152 $\pm$ 0.0014}   &   \textit{0.2396 $\pm$ 0.0019}  &  498.0s\\
    LINN                               &   \textbf{0.4064 $\pm$ 0.0015}*  &   \textbf{0.1850 $\pm$ 0.0053}*  &   30.7s
                                      &   \textbf{0.4191 $\pm$ 0.0012}*  &   \textbf{0.2438 $\pm$ 0.0014}*  &  754.9s\\
  \bottomrule
\end{tabular}
\begin{minipage}{12cm}
  \begin{tablenotes}
  \item * Significantly better than the best of other results (italic ones) with $p<0.05$
  \end{tablenotes}
\end{minipage}
\vspace{-3mm}
\end{table*}

\subsubsection{Baselines}

We compare LINN with the following baselines:

$\bullet$ \textbf{BPRMF}~\cite{rendle2009bpr}. This is a Matrix Factorization (MF)-based model that applies a pairwise ranking loss. It makes recommendations by the dot product result of user and item vectors plus biases, which is popular and competitive for item recommendation.

$\bullet$ \textbf{SVD++}~\cite{koren2008SVDPP}. This model integrates both explicit and implicit feedback, which explicitly models the user history interactions. It has been verified as a powerful model of personalized recommendation in many previous works.

Not only traditional recommendation models, but also some recent neural recommendation models are compared to LINN. In this work, we select the following neural models, which are more powerful compared with other neural matching-based models \cite{dacrema2019we}.



$\bullet$ \textbf{STAMP}~\cite{liu2018STAMP}. The Short-Term Attention/Memory Priority model, which uses the attention mechanism to model both short-term and long-term user preferences. This is one of the state-of-the-art recommendation models that consider the recent user interactions for recommendation.


$\bullet$ \textbf{GRU4Rec}~\cite{hidasi2015GRU4Rec}. A sequential recommendation model that applies Gated Recurrent Unit (GRU)~\cite{chung2014GRU} to derive the ranking scores.

$\bullet$ \textbf{NARM}~\cite{li2017NARM}. This model utilizes GRU and the attention mechanism to consider the importance of different interactions, which improves the performance of recommendation. It is a state-of-the-art sequential recommendation model. 

In our experiments, for LINN and all baseline models utilizing the sequential information (recent interactions), at most 10 previous interactions right before the target positive item are considered.

\subsubsection{Parameters}
All the models, including baselines, are trained with Adam~\cite{kingma2014adam} in mini-batches at the size of 128. The learning rate is 0.001 and early-stopping is conducted according to the performance on the validation set. Models are trained at most 100 epochs. To prevent models from overfitting, we use both the $\ell_2$-regularization and dropout. The weight of $\ell_2$-regularization $\lambda_\Theta$ is set between $1\times10^{-7}$ to $1\times10^{-4}$ and dropout ratio is set to 0.2. Vector sizes of the variables and the user/item vectors in the recommendation are 64. We run the experiments with five different random seeds and report the average results and standard errors. For LINN, $\lambda_{l}$ is set to $1\times10^{-6}$ on ML-100k and $1\times10^{-5}$ on Electronics. $\lambda_{\ell}$ is set to $1\times10^{-4}$ on ML-100k and $1\times10^{-6}$ on Electronics. Note that LINN has similar time and space complexity with baseline models, and each experiment run can be finished in 6 hours (several minutes on small datasets) with a single GPU (NVIDIA GeForce GTX 1080Ti).
\footnote{Codes are provided on GitHub at \url{https://github.com/rutgerswiselab/NLR}}

\vspace{-2ex}
\subsection{Overall Performance}
The overall performance of models on two datasets are shown in Table~\ref{tab:performance_rec}. From the results, 
STAMP achieves the best performance among the non-sequential baselines because it utilizes the attention mechanism to model both long-term and short-term user preferences. Models utilizing sequential information, such as GRU4Rec, and NARM, are significantly better than other non-sequential baselines, which is also observed by other works~\cite{hidasi2015GRU4Rec,li2017NARM}. It shows that sequential information is helpful to improve the recommendation performance. NARM performs better than GRU4Rec because it utilizes two attention networks to model the importance of interactions for users' local and global preferences. Although the personalized recommendation is not a standard logical inference problem, logical inference still helps in this task, which is shown by the results – we see that on both the ML100k and Electronics, LINN achieves the best performance. LINN makes more significant improvements on the Electronics dataset because the dataset is more sparse, and integrating logic inference is more beneficial than only using collaborative filtering. Note that without logic regularizers, LINN-$R_l$ still achieves comparable performance with NARM on ML-100k and is significantly better on Electronics. 

Our LINN model is simple in nature, which only learns three lightweight modules without personalized user embedding for all of the prediction tasks. It is interesting to see that such a simple approach outperforms state-of-the-art complex models such as NARM, which adopts a bi-attention-network and learns the user purpose features for personalization. This implies the power of reasoning based on modularized logic networks, and indicates a new way of designing neural networks and modeling structural data. We expect the performance of LINN will be further improved when adding advanced modeling techniques into the architecture, such as personalization, attention mechanism, and user's long/short-term preferences, which will be explored in the future.

We also show the training time per epoch of different models. Models explicitly modeling user history are generally slower than BPRMF, among which SVD++ takes more time for taking the full user history as inputs. Because of the dynamic computational graph, we simply pad meaningless tokens when forming the batches for LINN if the batch size is not fully filled, which causes some redundant computation (detailed implementation can be found in our code). Although LINN is slower than most of the baselines, they are at comparative levels, and the time cost is acceptable. Better implementations, such as parallelized optimization and reducing redundant computation (similar to GRU4Rec), may help improve the efficiency of LINN, which we will explore in the future.

\subsection{Weight of Logic Regularizers}

\begin{figure}[htbp]
\vspace{-2mm}
  \centering
  \includegraphics[width=1.0\linewidth,trim={0 0mm 0mm 0mm},clip]{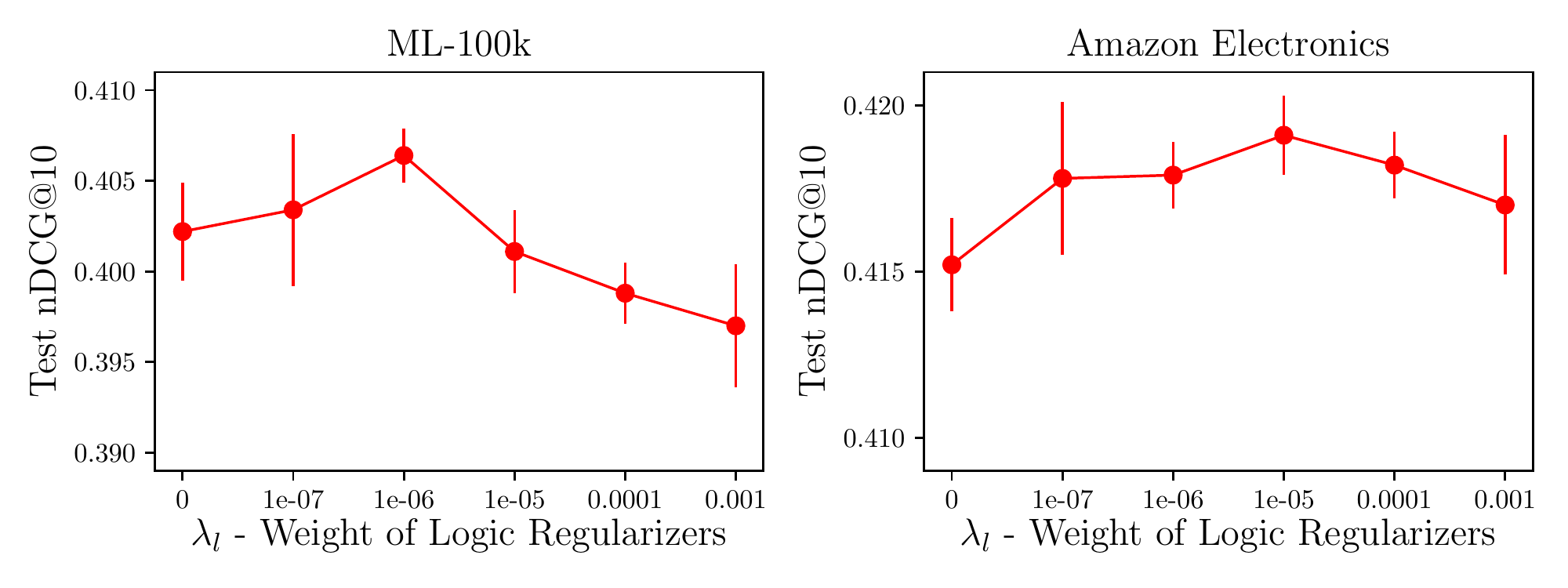}
  \vspace{-8mm}
  \caption{Performance on different logic regularizer weights.}
  \vspace{-5pt}
  \label{fig:rec_logic_weights}
\end{figure}

Results of using different weights of logical regularizers verify that logical inference is helpful in making recommendations, as shown in Figure~\ref{fig:rec_logic_weights}. Recommendation tasks can be considered as a logical inference task on the history of user behaviors, since certain user interactions may imply a high probability of interacting with another item.
On the other hand, learning the representations of users and items are more complicated than solving pure logical equations, since the model should have sufficient generalization ability to cope with noisy or even conflicting input expressions. Thus LINN, as an integration of logic inference and neural representation learning, performs well on the recommendation task. The weights of logical regularizers should not be too large because recommendation is not a pure logic reasoning problem but instead a reasoning problem on noisy data. Thus too large logical regularization weights may limit the expressiveness power and lead to a drop in performance.

\begin{table}[t!]
  \caption{Results about negative interactions in sequence}
   \vspace{-3mm}
  \label{tab:neg_interaction}
  \centering
  \begin{tabular}{l|ll|ll}
    \toprule
    \multicolumn{1}{c}{}&
    \multicolumn{2}{c}{\textbf{ML-100k}}&
    \multicolumn{2}{c}{\textbf{Amazon Electronics}}\\
    \cmidrule(lr){2-3} \cmidrule(lr){4-5}
    & nDCG@10 & Hit@1 & nDCG@10 & Hit@1\\
    \midrule
    \textbf{STAMP}                  &   0.3943   &   0.1706   &   0.3954   &   0.2215\\
    \textit{with} $-\mathbf{v}$     &   0.3757   &   0.1651   &   0.3901   &   0.2171\\
    \textit{with} $\mathbf{v}^-$    &   0.3803   &   0.1646   &   0.3936   &   0.2186\\
    \textit{with} $g(\mathbf{v})$   &   0.3745   &   0.1618   &   0.3899   &   0.2173\\
    \midrule
    \textbf{GRU4Rec}                &   0.3973   &   0.1745   &   0.4029   &   0.2262\\
    \textit{with} $-\mathbf{v}$     &   0.3909   &   0.1741   &   0.3924   &   0.2187\\
    \textit{with} $\mathbf{v}^-$    &   0.4006   &   0.1794   &   0.4050   &   0.2286\\
    \textit{with} $g(\mathbf{v})$   &   0.3843   &   0.1681   &   0.3946   &   0.2206\\
    \midrule
    \textbf{NARM}                   &   \textit{0.4022}   &   0.1771   &   \textit{0.4051}   &   \textit{0.2292}\\
    \textit{with} $-\mathbf{v}$     &   0.3917   &   0.1762   &   0.3978   &   0.2234\\
    \textit{with} $\mathbf{v}^-$    &   0.3964   &   \textit{0.1807}   &   0.4029   &   0.2262\\
    \textit{with} $g(\mathbf{v})$   &   0.3842   &   0.1743   &   0.3942   &   0.2206\\
    \midrule
    \textbf{LINN}                    &   \textbf{0.4064}*   &   \textbf{0.1850}*   &   \textbf{0.4191}*   &   \textbf{0.2438}*\\
    \textit{w/o negative}           &   0.3990   &   0.1822   &   0.4172   &   0.2433\\
  \bottomrule
\end{tabular}
\begin{minipage}{8cm}
  \begin{tablenotes}
  \item * Significantly better than the best baselines (italic ones) with $p<0.05$
  \end{tablenotes}
\end{minipage}
\vspace{-6mm}
\end{table}

\subsection{Negative Interactions in Sequences}

It should be noted that LINN can naturally model negative feedbacks in the user history through the negation operation $\neg$. As far as we know, except for using the negative interactions as the negative training samples, there is few work considering the negative interactions in user interaction sequences. Although considering negative interactions as negative samples helps to learn the item similarities and their relationships, removing them from user's historical sequence is not ideal for modeling user preference, because negative items indeed possess rich information about the user preference. LINN provides a straight-forward way to model the negative interactions in the sequence. For fair comparison, we enhance the STAMP, GRU4Rec, and NARM models in three ways so that they can also take sequences of both positive and negative interactions as inputs:
\begin{itemize}
    \item[(1)] Use the negative of the item vector $-\mathbf{v}$ instead of $\mathbf{v}$ as the input of RNN or attention networks in STAMP, GRU4Rec and NARM when the item is a negative interaction in the user's interaction sequence.
    \item[(2)] Build an extra item embedding matrix, so that each item has two vectors $\mathbf{v}$ and $\mathbf{v^-}$ for representing positive and negative interactions on the item, respectively.
    \item[(3)] When there comes a negative interaction, we first use a two-layer feed-forward network $g(\cdot)$ (similar as the NOT module in LINN) to transform the item vector $\mathbf{v}$ into its negative representation $g(\mathbf{v})$.
\end{itemize}

We try our best to tune the parameters, and the results are shown in Table~\ref{tab:neg_interaction}. 
We see that only GRU4Rec with an extra negative embedding matrix can slightly benefit from the negative interactions in the sequence. 
The reason may be that $-\mathbf{v}$ makes a too strong assumption that positive and negative preferences must have opposite representations in the vector space, while it may not be valid in non-linear neural networks and thus negatively affect the model performance. Using an unconstrained network $g(\cdot)$ only increases the parameter space and does not help in learning the meaning of \textit{negative} interactions. A separate embedding matrix for negative interactions is better than these two methods, but it also weakens the relationship between the two representations of the same item. However,
LINN achieves significantly better performance by using logic expressions to model the negative interactions in sequences. Note that even without these negative interactions, LINN still performs much better than GRU4Rec and NARM on Electronics, and achieves significantly better Hit@1 on ML-100k. 

\begin{figure}[t!]
  \centering
  \includegraphics[width=1.0\linewidth,trim={0 0mm 0 0mm},clip]{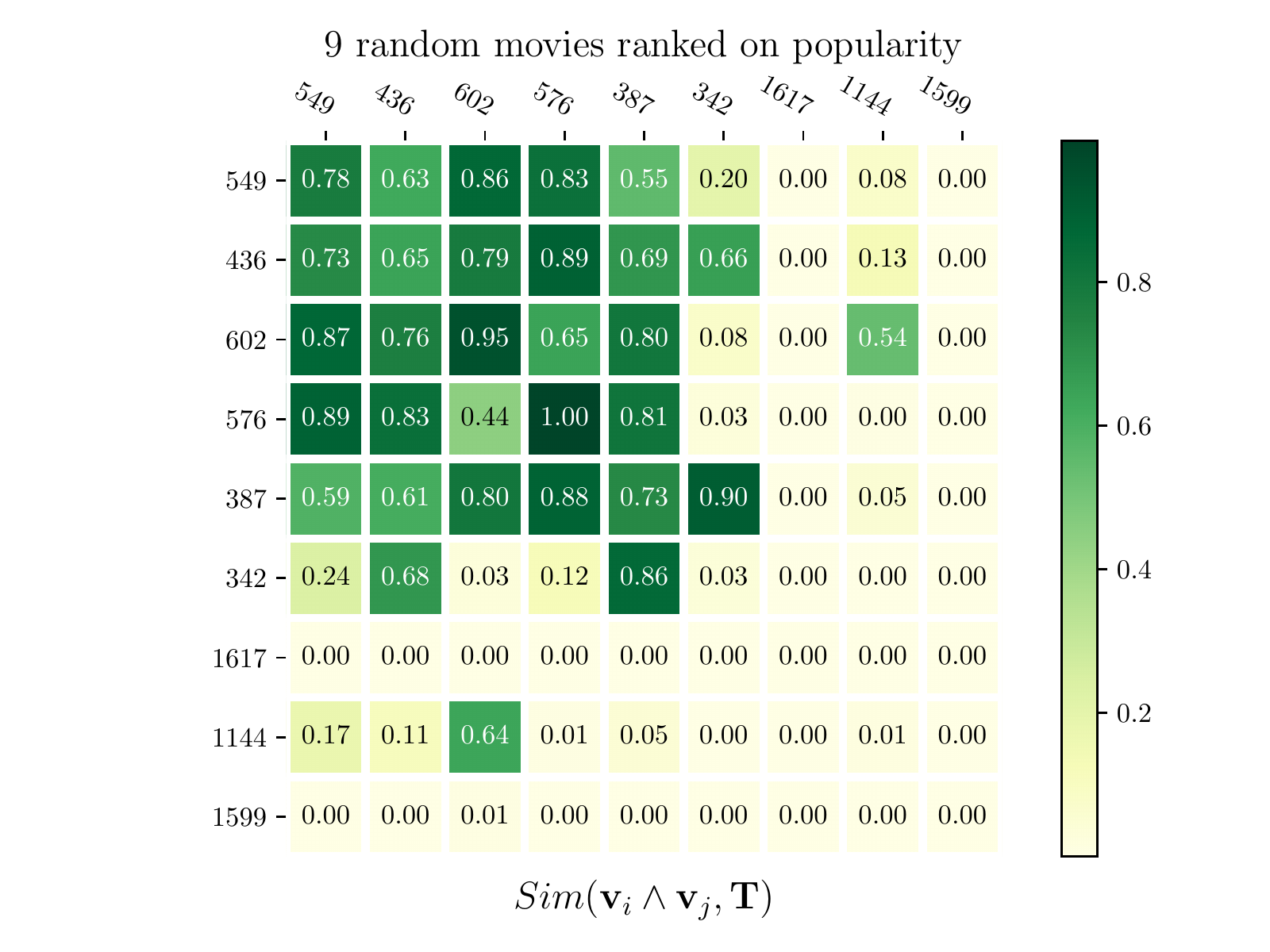}
  \vspace{-10mm}
  \caption{Heatmap of the co-occurrence probability by LINN.}
  \vspace{-20pt}
  \label{fig:show_matrix_sim}
\end{figure}

\subsection{Item Co-occurrence}

To better understand what LINN has learned, we randomly select 9 movies in ML-100k and use LINN to predict their co-occurrence (i.e., liked by the same user). The movies are ranked based on their popularity (interaction number) in the training set. More than 30 users interact with the first 3 movies, and less than 5 users interact with the last three. We use LINN to calculate the $Sim(\mathbf{v}_i\wedge \mathbf{v}_j, \mathbf{T})$ of each item pair, which means the probability of liking $v_j$ if a user likes $v_i$. The results are shown in Figure~\ref{fig:show_matrix_sim}, and we can draw the following conclusions:

\begin{itemize}
    \item Although we do not design a symmetric network for AND or OR, LINN successfully learns that $\mathbf{v}_i\wedge \mathbf{v}_j$ is close to $\mathbf{v}_j\wedge \mathbf{v}_i$. This is shown by the observation that symmetric positions in the heatmap matrix have similar values.
    \item LINN learns that popular item pairs have higher probabilities to be consumed by the same user, which corresponds to the top left corner of the heatmap matrix. 
    \item LINN can discover potential relationships between items. For example, item No.602 is a popular musical movie with song and dance, and item No.1144 is an unpopular drama movie, but they belong to the similar type of movies. We see that LINN is able to learn that they have a high probability of being liked by the same user.
    
\end{itemize}

Note that we do not use any content or category information of the movies to train the models. It is not easy for models to learn the item relationships solely based on interactions, especially on sparse data, while LINN does well based on the power of deep learning and logic reasoning. Our future work will consider modeling features and knowledge with logic-integrated neural networks, and we believe it has the potential to improve both transparency and effectiveness of deep neural networks.

\section{Conclusions and Future Work}
\label{sec:conclusion}
In this work, we propose a Logic-Integrated Neural Network (LINN) framework, which introduces logic reasoning into deep neural networks. 
In particular, we use latent vectors to represent the logic variables, and the logic operations are represented as neural modules regularized by logic rules.
The integration of logical inference and neural network enables the model to have the abilities of both representation learning and logical reasoning, which reveals a promising direction to design deep neural networks.
Experiments on theoretical task show that LINN works well on logic reasoning problems such as solving logical equations and variables. We further apply LINN to recommendation tasks effortlessly and achieve significant performance, which shows the prospect of LINN on practical tasks.

In this work, we focused on propositional logic reasoning with neural networks, while in the future, we will further explore predicate logic reasoning based on our LINN architecture, which can be easily extended by learning predicate operations as neural modules. We will also explore the possibility of encoding knowledge graph reasoning based on LINN, which will further contribute to the explainable AI research.

\section{Acknowledgement}
This work is supported in part by the Rutgers faculty support program, and in part by the National Key Research and Development Program of China (2018YFC0831900), Natural Science Foundation of China (61672311, 61532011), and Tsinghua University Guoqiang Research Institute. The project is also supported in part by China Postdoctoral Science Foundation and Dr. Weizhi Ma has been supported by Shuimu Tsinghua Scholar Program.

\bibliographystyle{ACM-Reference-Format}
\balance
\bibliography{reference.bib}


\end{document}